**Application of Neural Network in the Prediction of NOx Emissions from Degrading Gas Turbine**

Zhenkun Zheng and Alan Rezazadeh

Applied Research and Innovation Services

Southern Alberta Institute of Technology

1301 – 16 Avenue NW, Calgary, AB, Canada T2M 0L4

Zhenkun.Zheng@sait.ca

## ABSTRACT

This paper is aiming to apply neural network algorithm for predicting the process response (NOx emissions) from degrading natural gas turbines. Nine different process variables, or predictors, are considered in the predictive modelling. It is found out that the model trained by neural network algorithm should use part of recent data in the training and validation sets accounting for the impact of the system degradation. R-Square values of the training and validation sets demonstrate the validity of the model. The residue plot, without any clear pattern, shows the model is appropriate. The ranking of the importance of the process variables are demonstrated and the prediction profile confirms the significance of the process variables. The model trained by using neural network algorithm manifests the optimal settings of the process variables to reach the minimum value of NOx emissions from the degrading gas turbine system.

## INTRODUCTION

NOx, which represents Nitrogen Oxide (NO2) and Nitric Oxide (NO), are unfavorable byproducts from various industrial processes, such as gas turbine (Andrews, 2013; CORREA, 1993; Gokulakrishnan & Klassen, 2013; Okafor et al., 2020; Schorr & Chalfin, n.d.). Nitrogen Oxide is brown gas and can cause chronic lung disease if inhaled. Moreover, these gases can result into the formation of the acid rain, as well as damaging tropospheric ozone (Oluwoye et al., 2017). Therefore, it is very critical to predict and minimize the amount of NOx emissions based on the operation conditions or process variables (KAYA et al., 2019; Lewis, 2015; Pires et al., 2018).

Neural Network algorithm is an advanced and effective tool for developing an appropriate model for prediction in various applications (Li et al., 2021; Sakib et al., 2019; Zhang et al., 2019).

The collected dataset is normally divided into three set: training, validation, and test. The training set of data is used to train the model with the appropriate neuron structure. The obtained model is then validated by applying the validation set into the model and compare the predicted value with the known value. In the end, the validated model is tested with test data, which has not been used in the training or validation of the model, to measure the performance of that model. Neural network algorithm has been widely used in many fields such as autonomous driving, gaming, and language processing(Pham & Jeon, 2017; Sharma & Kaushik, 2017; Song et al., 2017). It has also been applied in the industry for the monitoring and prediction of the plant performance(Abba et al., 2020; Guo & Uhrig, 1992; Hamed et al., 2004; Hanbay et al., 2008). However, almost all equipment in any industrial plants is inevitably degrading overtime. The speed of degradation is mostly unpredictable and depends on many factors such as operation and environment conditions. For such a degrading system, it remains a critical issue of how to properly choose the dataset for training, validation, and test to attain an accurate model, which can later be used to obtain the optimal settings of the process variable for the desired process response value.

The main objective of this paper is to study the neural network algorithm as a tool in predicting NOx emissions from the gas turbine system with potential aging over time. Two models are developed in this study with different strategies of dividing dataset into training, validation, and test. The performances of these two models are compared, and it is found out that the proper way of allocating dataset into training, validation, and test for a degrading industrial system is to include the recent data into the training and validation. With the better model, the ranking of the importance of the process variables is demonstrated. The optimal settings of the process variables are revealed to achieve the lowest NOx emissions.

**DATASET AND METHODS**

The utilized dataset in this research has 10 columns and 36,733 rows (Rezazadeh, 2021). Each column represents one process variable or response. Each row represents one record at a specified timestamp (every hour from year 2011 to 2015). In addition to the process response of the NOx emissions, process variables or predictors are as follow:

- Ambient temperature (AT)
- Ambient pressure (AP)
- Ambient humidity (AH)

- Air filter difference pressure (AFDP)
- Gas turbine exhaust pressure (TEP)
- Turbine inlet temperature (TIT)
- Turbine exhaust temperature (TET)
- Compressor discharge pressure (CDP)
- Turbine energy yield (TEY)

In this study, neural network algorithm is applied to predict the process response variable, NOx values from all the above nine predictors. The neural network algorithm contains two layers of neurons. Each layer has three TanH activation functions, one linear function, and one Gaussian function considering the linear correlation and nonlinear relationship in the dataset.

**RESULTS AND DISCUSSION**

Table 1 shows the histogram and boxplot of process response (NOx emissions) and process variables for the dataset ranging from year 2011 to 2015. Although some data points are located outside the whiskers of the boxplot, i.e., outside 1.5 times the interquartile range above the upper quartile or below the lower quartile, these data points are not due to the process upsets and therefore determined to be valid and reasonable. They are all included in further analyses.

*Table 1. Histogram and Boxplot of Process Response and Process Variables*

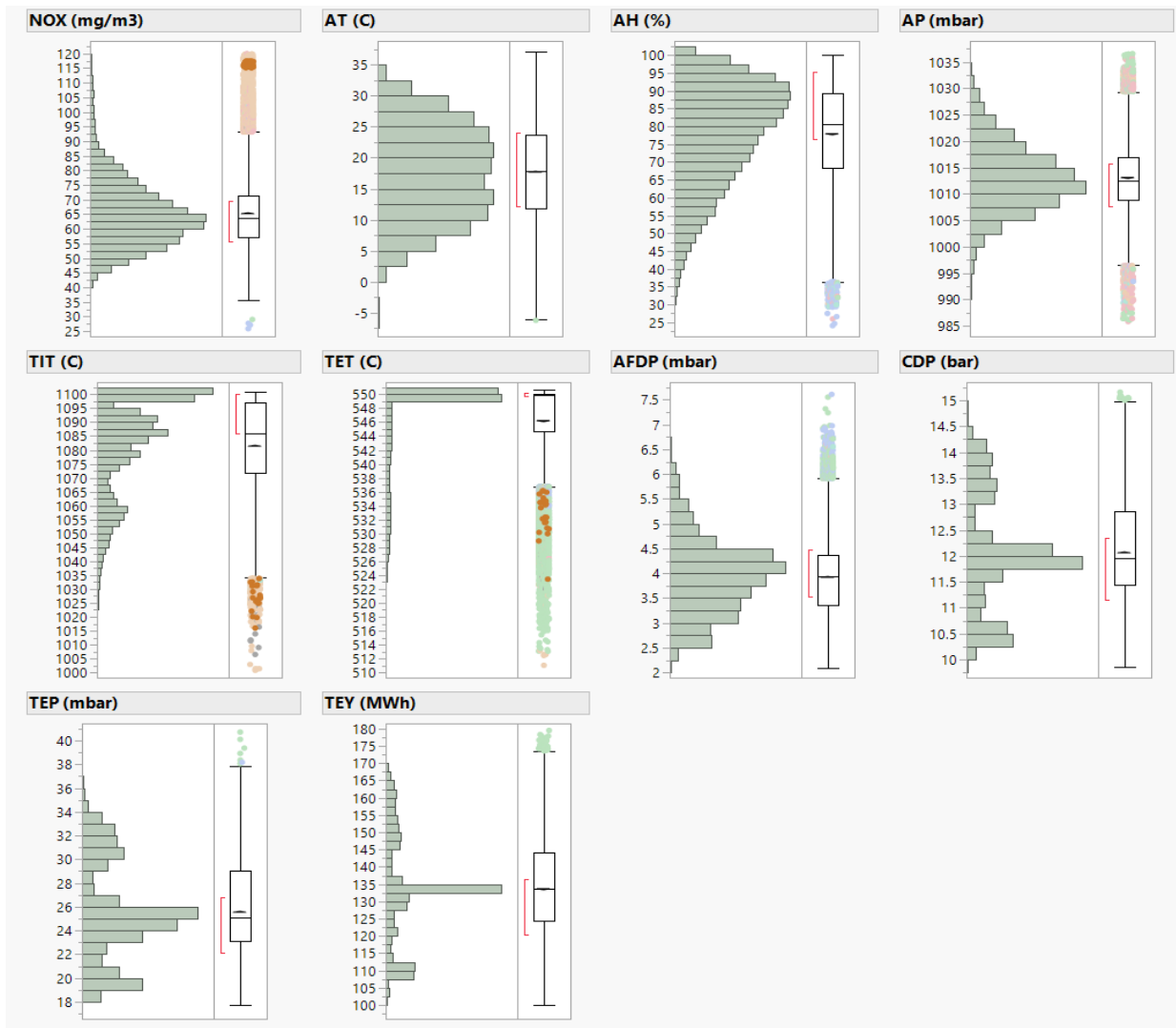

Table 2 illustrates the correlation analysis among process variables and process response. As can be seen in Table 2, TEP, TIT, CDP and TEY have very strong positive correlations between each other (i.e., correlation coefficient larger than 0.8). It is indicated that turbine energy yield is dependent on the turbine inlet temperature, compressor discharge pressure, and turbine exhaust pressure.

*Table 2. Correlation Analysis of Process Response and Process variables*

|  | NOX (mg/m3) | AT (C) | AH (%) | AP (mbar) | TIT (C) | TET (C) | AFDP (mbar) | CDP (bar) | TEP (mbar) | TEY (MWh) |
|---|---|---|---|---|---|---|---|---|---|---|
| NOX (mg/m3) | 1.0000 | -0.5582 | 0.1646 | 0.1919 | -0.2139 | -0.0928 | -0.1882 | -0.1713 | -0.2016 | -0.1161 |
| AT (C) | -0.5582 | 1.0000 | -0.4763 | -0.4066 | 0.1837 | 0.2819 | 0.2520 | 0.0153 | 0.0459 | -0.0912 |
| AH (%) | 0.1646 | -0.4763 | 1.0000 | -0.0152 | -0.2218 | 0.0230 | -0.1478 | -0.1963 | -0.2352 | -0.1374 |
| AP (mbar) | 0.1919 | -0.4066 | -0.0152 | 1.0000 | -0.0054 | -0.2256 | -0.0404 | 0.1026 | 0.0575 | 0.1182 |
| TIT (C) | -0.2139 | 0.1837 | -0.2218 | -0.0054 | 1.0000 | -0.3809 | 0.6913 | 0.9085 | 0.8742 | 0.9103 |
| TET (C) | -0.0928 | 0.2819 | 0.0230 | -0.2256 | -0.3809 | 1.0000 | -0.4669 | -0.7064 | -0.6997 | -0.6824 |
| AFDP (mbar) | -0.1882 | 0.2520 | -0.1478 | -0.0404 | 0.6913 | -0.4669 | 1.0000 | 0.7026 | 0.6785 | 0.6655 |
| CDP (bar) | -0.1713 | 0.0153 | -0.1963 | 0.1026 | 0.9085 | -0.7064 | 0.7026 | 1.0000 | 0.9785 | 0.9888 |
| TEP (mbar) | -0.2016 | 0.0459 | -0.2352 | 0.0575 | 0.8742 | -0.6997 | 0.6785 | 0.9785 | 1.0000 | 0.9641 |
| TEY (MWh) | -0.1161 | -0.0912 | -0.1374 | 0.1182 | 0.9103 | -0.6824 | 0.6655 | 0.9888 | 0.9641 | 1.0000 |

Neural network algorithm is used to predict the process response variable, NOx values. This technique is specifically advantageous for process variables with strong correlations. Two layers of Neurons are used in this research. Each layer has three TanH activation functions, one linear function, and one Gaussian function considering the linear correlation and nonlinear relationship in the dataset. The first model uses data from year 2011 to 2013 as training set, data from year 2014 as validation set, and data from year 2015 as test set. As shown in table 3, R-Square of the training set is 0.72 and R-Square of the validation set is 0.35. It is indicated that the trained model can not be well validated by using the following year data. R-Square of test set further dropped to 0.05 when using data from year 2015 as the test set. The poor performance of the model in predictions can be explained by the severe degradation of the gas turbine system over time The trained model only account for the system condition for year 2011-2013 and can not predict the future performance due to the system aging.

*Table 3. Training, Validation, and Test Results of the Neural Network Trained with Y2011-2013 data, Validated with Y2014 data, and tested with Y2015 data*

| Model NTanH(3)NLinear(1)NGaussian(1)NTanH2(3)NLinear2(1)NGaussian2(1) | | | | | |
|---|---|---|---|---|---|
| **Training** | | **Validation** | | **Test** | |
| NOX (mg/m3) | | NOX (mg/m3) | | NOX (mg/m3) | |
| **Measures** | **Value** | **Measures** | **Value** | **Measures** | **Value** |
| RSquare | 0.7188445 | RSquare | 0.3517042 | RSquare | 0.0536235 |
| RMSE | 5.8517301 | RMSE | 8.0283585 | RMSE | 10.829137 |
| Mean Abs Dev | 4.0973904 | Mean Abs Dev | 6.2606947 | Mean Abs Dev | 9.5568066 |
| -LogLikelihood | 70693.334 | -LogLikelihood | 25066.733 | -LogLikelihood | 28067.905 |
| SSE | 759880.77 | SSE | 461365.6 | SSE | 865923.25 |
| Sum Freq | 22191 | Sum Freq | 7158 | Sum Freq | 7384 |

To account for the system degradation in the model, the dataset is stratified by year with 60% for training set, 20% for validation set, and 20% for test set. In this method, the trained model will include the degradation effect since training set covers year 2011-2015. The performance of the trained model has improved significantly as shown in Table 4. R-Squares of the training, validation, and test are 0.82,0.81, and 0.82, respectively. Moreover, R-Square of the test set is closer to that of the training and validation set, suggesting that the model is not overfitted or underfitted. Therefore, the second model is adopted and further investigated in this research.

*Table 4. Training, Validation, and Test Results of the Neural Network Trained with Dataset Stratified by Year for 60% data(training), 20%(validation), and 20%(test)*

| Model NTanH(3)NLinear(1)NGaussian(1)NTanH2(3)NLinear2(1)NGaussian2(1) | | | | | |
|---|---|---|---|---|---|
| **Training** | | **Validation** | | **Test** | |
| **NOX (mg/m3)** | | **NOX (mg/m3)** | | **NOX (mg/m3)** | |
| **Measures** | **Value** | **Measures** | **Value** | **Measures** | **Value** |
| RSquare | 0.824835 | RSquare | 0.8130671 | RSquare | 0.8217441 |
| RMSE | 4.8593385 | RMSE | 5.0116331 | RMSE | 5.0508789 |
| Mean Abs Dev | 3.4078844 | Mean Abs Dev | 3.4236446 | Mean Abs Dev | 3.4933248 |
| -LogLikelihood | 66116.492 | -LogLikelihood | 22266.556 | -LogLikelihood | 22320.827 |
| SSE | 520434.28 | SSE | 184530.68 | SSE | 187406.58 |
| Sum Freq | 22040 | Sum Freq | 7347 | Sum Freq | 7346 |

.

Figure 1 confirms the accuracy of the second model trained by neural network algorithm. In statistics, the actual value is the value that is obtained by measurement. The predicted value is the value of the variable predicted based on the model. Ideally the actual and predicted values should form a diagonal line in the plot. In this research, the actual vs predicted plot of the obtained model is very accurate since the data scattered symmetrically about a 45-degree diagonal.

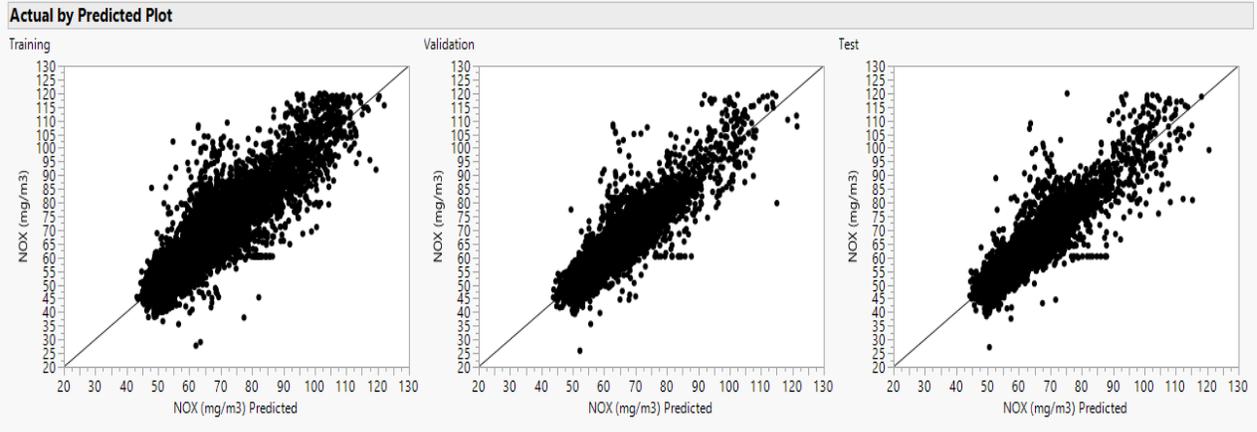

*Figure 1. Plot of Actual vs Predicted of Process Response (NOx)*

A residual plot is a scatterplot that displays the residuals on the vertical axis and the independent variable on the horizontal axis. Residual plots shown in Fig. 2 further confirm that the model is appropriate since not any clear pattern is observed from the residue plot.

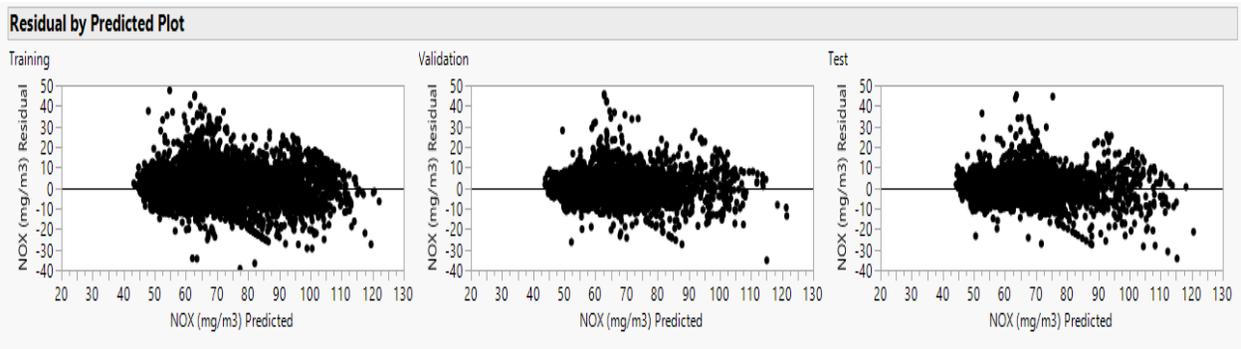

*Figure 2. Residue Plot of Process Response (NOx)*

It is obvious that nine process variables studied in this project have different impacts on the process response (NOx). Table 4 summaries the ranking of the importance of process variables in the model. Turbine inlet temperature (TIT) is of the most importance in the prediction of the process response followed by TEY, TEP, TET, AT, and CDP. AH, AP, and AFDP has very little effect on the process response.

*Table 4. Ranking of the process variable importance for the prediction of process response (NOx)*

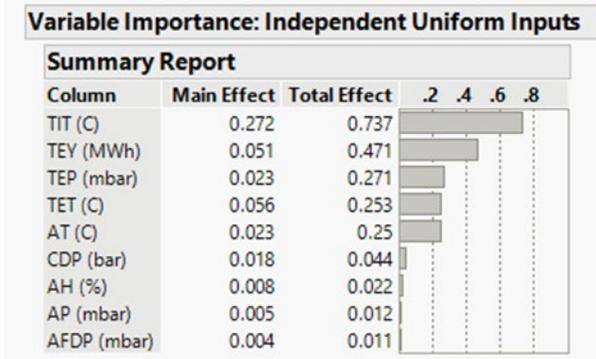

The prediction profiler shows the predicted response at specified values of each of the process variables, which are listed across the bottom in Fig. 3. The importance of different process variables on the prediction of process response can be also shown in the prediction profile. As can be seen in Fig.3, No significant change is observed on the process response with different settings of AH, AP, and AFDP since the corresponding curve are almost horizontal. In desirability profiling, the desirability function can be specified for the process response. In this research, the lowest value of the process response (NOx) is most desired. Fig. 3 shows the values of each of the process variable in order for the gas turbine plant to minimize the NOx emission at ~22.3 mg/m³. The prediction profile provides valuables insights into the optimal operation of gas turbines.

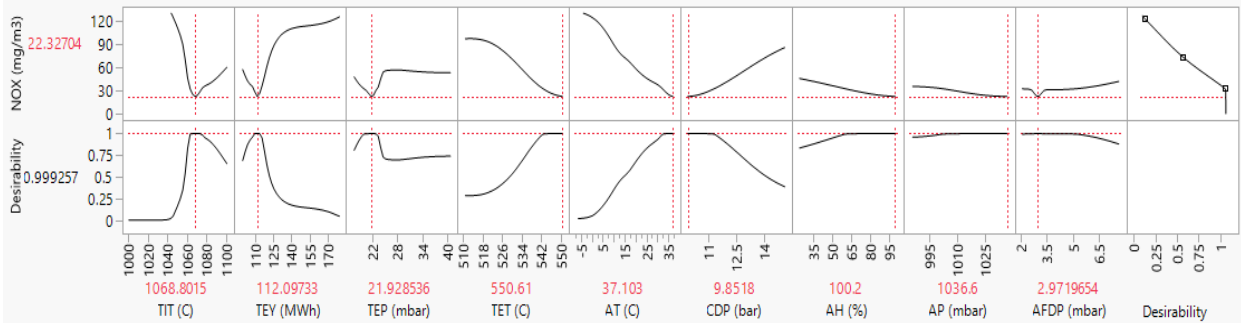

*Figure 1. Prediction profile of process response NOx with different settings of process variables*

.

## CONCLUSION

Neural network algorithm is an effective technique in the prediction of NOx emissions from the gas turbine. R-Squares of the training set and validation set are both above 0.8. This technique not only shows the rank of the importance of the process variables, but also reveals the required settings of process variables to obtain desired process response. This outcome provides significant insights for the gas turbine operation.

## ACKNOWLEDGEMENT

The dataset used in this analysis was produced by a power generation plant in Turkey. This research could not be possible without the shared data with the scientific community. Author would like to express gratitude to the power generator operator for publishing and sharing the operational data.

Author would like to acknowledge CFREF (Canada First Research Excellence Fund) for providing the opportunity to conduct this research.

## GRANT SUPPORT DETAILS

The present research did not receive any financial support.

## CONFLICT OF INTEREST

The author declares that there is not any conflict of interests regarding the publication of this manuscript. In addition, the ethical issues, including plagiarism, informed consent, misconduct, data fabrication and/or falsification, double publication and/or submission, and redundancy has been completely observed by the author.

## LIFE SCIENCE REPORTING

No life science threat was practiced in this research.